\documentclass[letterpaper, 10 pt, conference]{ieeeconf}  % Comment this line out if you need a4paper
\IEEEoverridecommandlockouts                              % This command is only needed if
\usepackage[T1]{fontenc}
\usepackage[noadjust]{cite}

\usepackage{amsmath}
\usepackage{yhmath} %note this package will generate type 3 fonts unless you have 1.6 installed
\usepackage{mathtools} % for rotation matrix e.g. \prescript{4}{0}{R} and for summation
\usepackage{amssymb} %for trianleq symbol
\usepackage{sidecap}
\usepackage{graphicx}
\usepackage{subcaption}
\usepackage{multirow}
\usepackage{caption}
\usepackage[super]{nth} %to type 1st,2nd,3rd,... , use the command \nth {<number>}
\usepackage[normalem]{ulem} % to strike out
\usepackage{color}
\usepackage{xcolor} %for coloring text
\usepackage{booktabs} %nice tables
\usepackage{siunitx} %e.g. \ang{45} to show 90 degree
\usepackage{booktabs} %nice tables
\usepackage{colortbl} % to color tables
\usepackage{textcomp} %for a nice registered trademark, e.g. MATLAB\textsuperscript{\textregistered}
\usepackage{tikz}
\usepackage[noadjust]{cite} %to group citations

\usepackage{wasysym}  % for diameter symbol e.g. \diameter

\usepackage{color}
\usepackage{hyperref}  %this package hyperrefs to figure and equation numbers. see http://www.tug.org/applications/hyperref/manual.html
\definecolor{darkblue}{rgb}{0.0,0.0,1}
\hypersetup{colorlinks=false,breaklinks,
            linkcolor=darkblue,urlcolor=darkblue,
            anchorcolor=darkblue,citecolor=darkblue}

\newcommand{\RNum}[1]{\uppercase\expandafter{\romannumeral #1\relax}} %Roman numerals

\newcommand{\realfield}[1]{\hbox{I \kern -.5em R}^{#1}}
\newcommand {\mb}[1]{\mathbf{#1}}
\newcommand {\bs}[1]{\boldsymbol{#1}}

\newcommand{\T}{^{\mathrm{T}}}  %shortcut for transpose

\definecolor{LightGray}{gray}{0.9}
\newcolumntype{a}{>{\columncolor{LightGray}}l}
\newcommand{\PreserveBackslash}[1]{\let\temp=\\#1\let\\=\temp}
\newcolumntype{C}[1]{>{\PreserveBackslash\centering}p{#1}}
\newcolumntype{L}[1]{>{\PreserveBackslash}l{#1}}

\usepackage{fancyhdr} 
\usepackage{color}  %needed for color text

\usepackage[textsize=scriptsize, bordercolor=white,backgroundcolor=gray!30,linecolor=black,colorinlistoftodos]{todonotes}  %for adding comments. See http://www.texample.net/tikz/examples/todo-notes/.
\usepackage[normalem]{ulem} % added for strikthrough text use   \sout{text you want to strike through
\usepackage{soul} %added for \texthl for the highlight and also for \ul for underline across wrapped text

\title{\LARGE \bf
Solving Cosserat Rod Models via \\ Collocation and the Magnus Expansion
}
\author{~Andrew~L.~Orekhov$^{1}$,~Nabil~Simaan$^{1}$$^{\dag}$ % <-this % stops a space
\thanks{$\dag$ Corresponding author}
\thanks{$^{1}$Department of Mechanical Engineering, Vanderbilt University, Nashville, TN 37235, USA
        {\tt (andrew.orekhov, nabil.simaan) @vanderbilt.edu}}
\thanks{This work was supported by NSF award \#1734461 and by Vanderbilt University funds. A. Orekhov was partly supported by the NSF Graduate Research Fellowship under \#DGE-1445197. }
}

\usepackage{balance}
\begin{document}
\maketitle
\thispagestyle{empty}
%%%%%%======== FOR ARXIV copyright footer note ==================
%%  Uncomment this section to add header and footer with copyright
%==============================================================
\thispagestyle{fancy}
\fancyhf{}
\renewcommand{\headrulewidth}{0pt}
\lhead{2020 IEEE/RSJ International Conference on Intelligent Robots and Systems (IROS). Accepted Version. }
\rfoot{\centering \scriptsize \copyright 2020 IEEE. Personal use of this material is permitted. Permission from IEEE must be obtained for all other uses, in any current or future media, including reprinting/republishing this material for advertising or promotional purposes, creating new collective works, for resale or redistribution to servers or lists, or reuse of any copyrighted component of this work in other works.}
%==============================================================
\pagestyle{empty}
%timer
\begin{abstract}
Choosing a kinematic model for a continuum robot typically involves making a tradeoff between accuracy and computational complexity. One common modeling approach is to use the Cosserat rod equations, which have been shown to be accurate for many types of continuum robots. This approach, however, still presents significant computational cost, particularly when many Cosserat rods are coupled via kinematic constraints. In this work, we propose a numerical method that combines orthogonal collocation on the local rod curvature and forward integration of the Cosserat rod kinematic equations via the Magnus expansion, allowing the equilibrium shape to be written as a product of matrix exponentials. We provide a bound on the maximum step size to guarantee convergence of the Magnus expansion for the case of Cosserat rods, compare in simulation against other approaches, and demonstrate the tradeoffs between speed and accuracy for the fourth and sixth order Magnus expansions as well as for different numbers of collocation points. Our results show that the proposed method can find accurate solutions to the Cosserat rod equations and can potentially be competitive in computation speed.
\end{abstract} 
%===========================================================
%============================================================
\section{Introduction} \label{sec:intro}

\par Continuum and soft robot architectures have been studied for a variety of useful applications \cite{burgner2015continuum,rus2015design}, but their passive, continuously flexible structures make them difficult to model. There are many different kinematic and dynamic models presented in the literature (see \cite{burgner2015continuum,rus2015design,webster_design_2010,sadati_tmtdyn_2020} for reviews), but one commonly used method is to model the robot's flexible structure as one or more Cosserat rods. This approach has been experimentally validated for concentric tube robots \cite{dupont2009design,rucker_geometrically_2010}, tendon-driven robots \cite{rucker_statics_2011,chikhaoui_comparison_2019}, multi-backbone robots \cite{orekhov_modeling_2017}, and soft robots with fluidic \cite{sadati_control_2018} and tendon \cite{renda_screw-based_2017} actuation in both kinematic and dynamic studies \cite{till2019real}.
\par Experimental validations have shown relatively accurate open-loop prediction of shape (position errors of 1-8\% of arc length are typical), however, computing the model involves numerically solving a set of boundary value problems (BVPs) which can be computationally expensive. Although a number of works have demonstrated implementations fast enough for control in the cases of concentric tube robots \cite{dupont2009design}, parallel continuum robots \cite{till_efficient_2015}, and single-backbone tendon-driven robots \cite{till2019real}, less accurate modeling methods are still attractive due to their low computational cost when compared to the Cosserat rod models \cite{chikhaoui_comparison_2019}. Furthermore, in cases where the model consists of many kinematically coupled Cosserat rods (e.g. multi-backbone robots \cite{simaan2005snake,orekhov_modeling_2017} and eccentric pre-curved tube robots \cite{wang2019steering}), the computational cost of the Cosserat rod models is a significant obstacle and more efficient numerical methods are still needed.
\par Another drawback of Cosserat rod modeling is that after the BVP has been numerically solved (typically with a shooting method), it can be difficult to compute forward/inverse Jacobians when many kinematic constraints are active. Although in many cases the partial derivatives associated with the Jacobian can be computed together with the forward integration of the Cosserat differential equations (DEs) \cite{rucker2011computing}, for more complicated continuum structures one typically has to resort to finite-difference estimation. A numerical method that results in an analytical expression can aid in computing these matrices and conducting other analytical analysis of the rod equilibrium shape for design and control.
\par This paper is motivated by these two limitations of Cosserat rod models and is a preliminary step towards addressing them. We propose a method that solves for the rod's curvature distribution with global orthogonal collocation and uses the Magnus expansion, a Lie group integration method, to recover the shape of the rod. Solving the BVP in this way provides some computational advantages and results in a product of matrix exponentials expression for the shape, which, as shown in \cite{renda_discrete_2018,renda_geometric_2020}, allows the Jacobian to be computed in closed-form. Although we consider the case of a single rod and additional evaluation is need for practical robotics scenarios, our results show that this method can potentially be competitive with other approaches.
\par Collocation, and other weighted residuals methods, have been demonstrated previously \cite{sadati_control_2018,sadati_tmtdyn_2020,weeger2017isogeometric,liu_spectral_2018,khaneh_masjedi_chebyshev_2015}, but these works apply polynomial interpolation on internal wrenches or on the position and orientation instead of curvature. Curvature-based parameterizations have been used in \cite{renda_discrete_2018,grazioso_geometrically_2018}, but they do not combine this with collocation, and they use constant-twist deformation elements, in contrast to the global interpolation functions we use here. A polynomial curvature model was used in \cite{santina_control_2020}, although this work was focused on control of a planar robot and not modeling of Cosserat rods. In \cite{vcevsarek2013dynamics}, a curvature parameterization is combined with collocation for Cosserat rod dynamics, but the Magnus expansion is not used.
\par Our method is most similar to recent works \cite{renda_geometric_2020,boyer2019forward} that also use polynomial approximations on curvature within a Lie-group framework. The Magnus expansion is used in \cite{renda_geometric_2020} and Chebyshev polynomials are used in \cite{boyer2019forward}. This paper complements and extends these works by 1) demonstrating the computational benefits of combining Chebyshev orthogonal collocation with the Magnus expansion for Cosserat rod BVPs, 2) analysing the maximum step sizes to guarantee convergence of the Magnus expansion, and 3) validating the approach with high-order polynomials against two known methods in a simulation study.
%===============================================================================
\section{The Cosserat Rod Equations}
Here we briefly review the Cosserat rod equations of static equilibrium. We assume that shear strains and extension are negligible, which has been shown to be a reasonable assumption for long, slender rods \cite{dupont2009design}. We also assume the rod is not subject to distributed loads, is not pre-curved, and has a uniform cross-section and bending stiffness. Although we consider this simpler case for brevity,  the methods in this paper extend to these more general cases, which have been discussed in other work \cite{rucker_statics_2011}.
\begin{figure}[htbp]\center
\includegraphics[width=0.99\columnwidth]{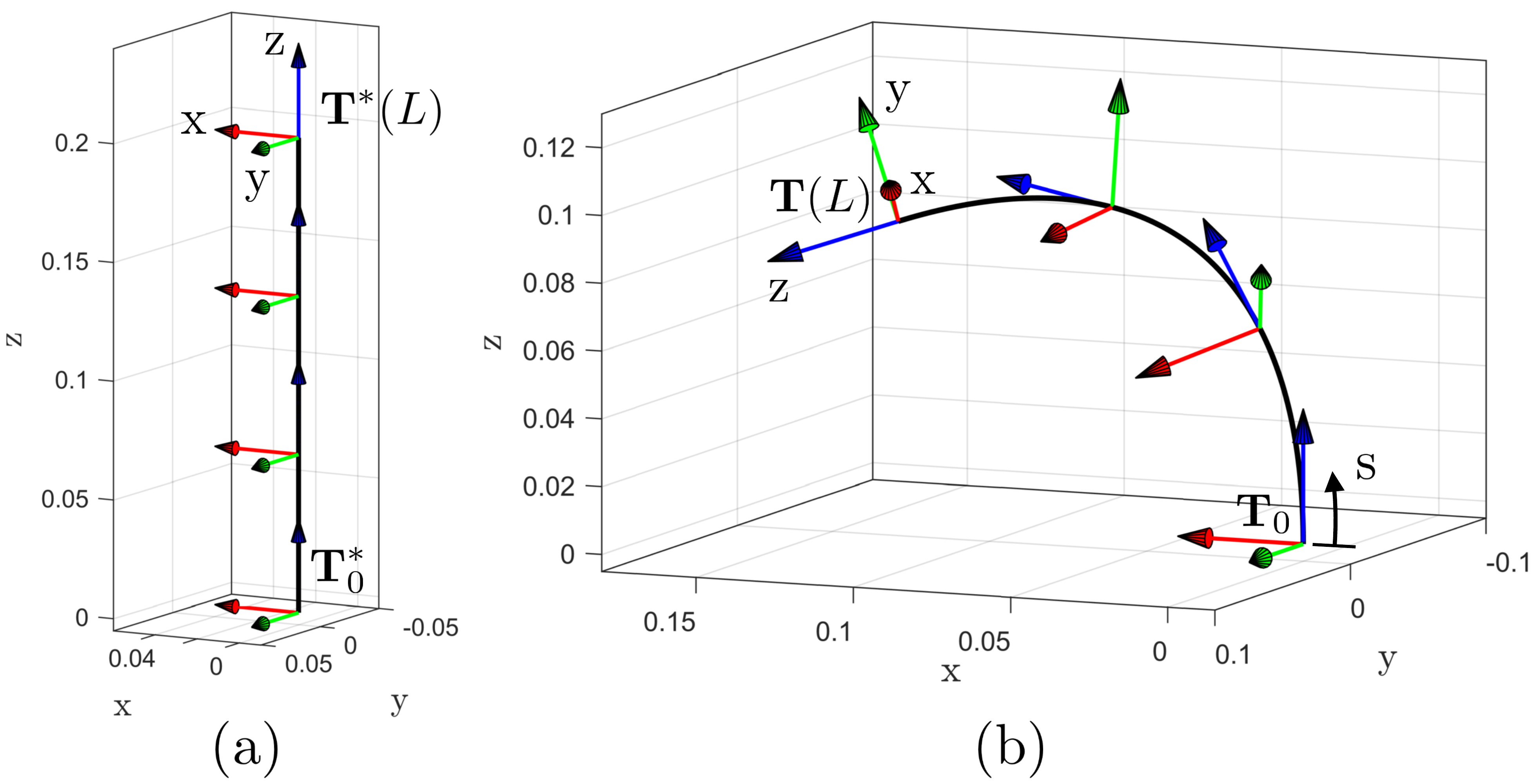}
\caption{Kinematic notation and frame assignment: (a) the rod in its undeflected state with reference frames $\mb{T}^*(s)$, and (b) the rod after undergoing a spatial deflection.}
\label{fig:framed_curve}
\end{figure}
\par Under these assumptions, the shape of a deflected rod of length $L$ is parameterized as a matrix function of arc-length $s \in [0, L]$ given as a homogenous transformation $\mb{T}(s)$:
\begin{equation}
\begin{aligned}
\mb{T}(s) = \begin{pmatrix}\mb{R}(s) & \mb{p}(s) \\ 0 & 1\end{pmatrix} \in SE(3)\\
\end{aligned}
\end{equation}
 where $\mb{p}(s) \in \realfield{3}$ specifies the rod's shape in 3D and $\mb{R}(s) \in SO(3)$ describes the orientation of each local material frame (as shown in Fig. \ref{fig:framed_curve}).
\par We assign reference frames $\mb{T}^*(s)$ to the unloaded rod's curve as shown in Fig. \ref{fig:framed_curve}(a) with the $z$-axis of $\mb{T}^*$ aligned with the rod and pointing towards its tip.
Defining $\mb{u} = [u_x,u_y,u_z]\T$ as a vector of curvatures, we describe the motion of $\mb{T}(s)$ along the curve for unit-speed traversal along the arc-length by a twist vector $\bs{\xi} = [\mb{u}\T(s),0,0,1]\T$ expressed in the moving frame. Using the wedge operator $ ^\wedge $, we map $\mb{u}$ to its skew-symmetric matrix $\widehat{\mb{u}} \in so(3)$ and $\bs{\xi}$ to its $se(3)$ element, which is defined as:
\begin{equation} \label{eq:Tdot_def}
\begin{aligned}
\widehat{\bs{\xi}}(s) = \mb{X}(s) &=\begin{pmatrix} \widehat{\mb{u}}(s) & \mb{e}_3 \\ 0 & 0\end{pmatrix} \in se(3)\\
 \mb{X}(s) &= \mb{T}^{-1}(s)\mb{T}'(s)
\end{aligned}
\end{equation}
where $\mb{T}'(s)$ is the derivative of $\mb{T}$ with respect to arc-length and $\mb{e}_3 = [0,0,1]\T$. We use a moving frame twist so that the rod's internal moment can be obtained directly from the local curvature:
\begin{equation}
\mb{m}(s) = \mb{R}(s)\mb{K}\mb{u}(s), \quad \mb{m}(s) \in \realfield{3}
\end{equation}
where $\mb{K} = \text{diag}(EI,EI,JG)$ is the rod's arc-length normalized bending stiffness matrix. Note we have expressed $\mb{m}(s)$ in world frame since external wrenches are more easily expressed in world frame.
\par For a known tip-applied wrench, the Cosserat rod ordinary differential equations (ODEs) simplify to \cite{rucker_statics_2011}:
\begin{equation}  \label{eq:cosserat_ode}
\begin{aligned}
\mb{T}'(s) &= \mb{T}(s)\mb{X}(s) \quad \\
\mb{u}'(s) &= \mb{g}(\mb{u}(s)) = -\mb{K}^{-1}\left( \widehat{\mb{u}}(s)\mb{K}\mb{u}(s) + \widehat{\mb{e}}_3\mb{R}\T(s)\mb{f}_e \right)
\end{aligned}
\end{equation}
where $\mb{u}'(s)$ denotes a derivative with respect to $s$ and $\mb{f}_e$ is a force at the rod's tip expressed in world frame. The boundary conditions for a known applied tip wrench are given by:
\begin{equation} \label{eq:boundary_condition}
\mb{u}(L) = \mb{K}^{-1}\mb{R}\T(L)\mb{m}_e
\end{equation}
where $\mb{m}_e = [m_{e,x},m_{e,y},m_{e,z}]\T$ is the tip moment expressed in world frame. Solving the ODE's in (\ref{eq:cosserat_ode}) with the boundary conditions (\ref{eq:boundary_condition}) provides the frames $\mb{T}(s)$ along the rod which gives the shape of the rod in space.
\par This boundary value problem (BVP) is typically solved via the shooting method. This has been successfully applied to a number of continuum robot architectures for forward/inverse kinematics \cite{black2017parallel} and forward dynamics \cite{till2019real}. One drawback of the shooting method is that it involves many forward integrations of $\mb{T}'(s)$ and $\mb{u}'(s)$ and therefore can be computationally expensive for Cosserat rod models with many kinematic constraints (e.g. \cite{orekhov_modeling_2017,wang2019steering}). Another drawback is that, once a solution to the BVP is found, computing Jacobians requires finite difference approximation. Several works addressed this issue for some continuum robot architectures \cite{rucker2011computing,till_efficient_2015}, but this difficulty remains for architectures with many coupled Cosserat rods.
\par In the sections below, we describe a procedure for solving (\ref{eq:cosserat_ode}) and (\ref{eq:boundary_condition}) that uses collocation to solve $\mb{u}'(s)$, which circumvents the need to numerically integrate $\mb{u}'(s)$, and a result from the geometric integration literature called the \emph{Magnus expansion} to forward integrate $\mb{T}'(s)$. The combination of these two known techniques allows for the solution to the BVP to be expressed as a product of matrix exponentials, an analytical expression that can be used for further analytical analysis (e.g. Jacobian-based studies \cite{renda_discrete_2018,grazioso_geometrically_2018}).
%

%===============================================================================
\section{Solving via Orthogonal Collocation}
\par Here we show how to solve (\ref{eq:cosserat_ode}) via orthogonal collocation, a direct variational approach that has been applied to a variety of DEs and BVPs \cite{young2019orthogonal,canuto2007spectral,mason2002chebyshev}. First, we review polynomial interpolation and collocation.
\begin{figure}[h]
\center
\includegraphics[width=0.88\columnwidth]{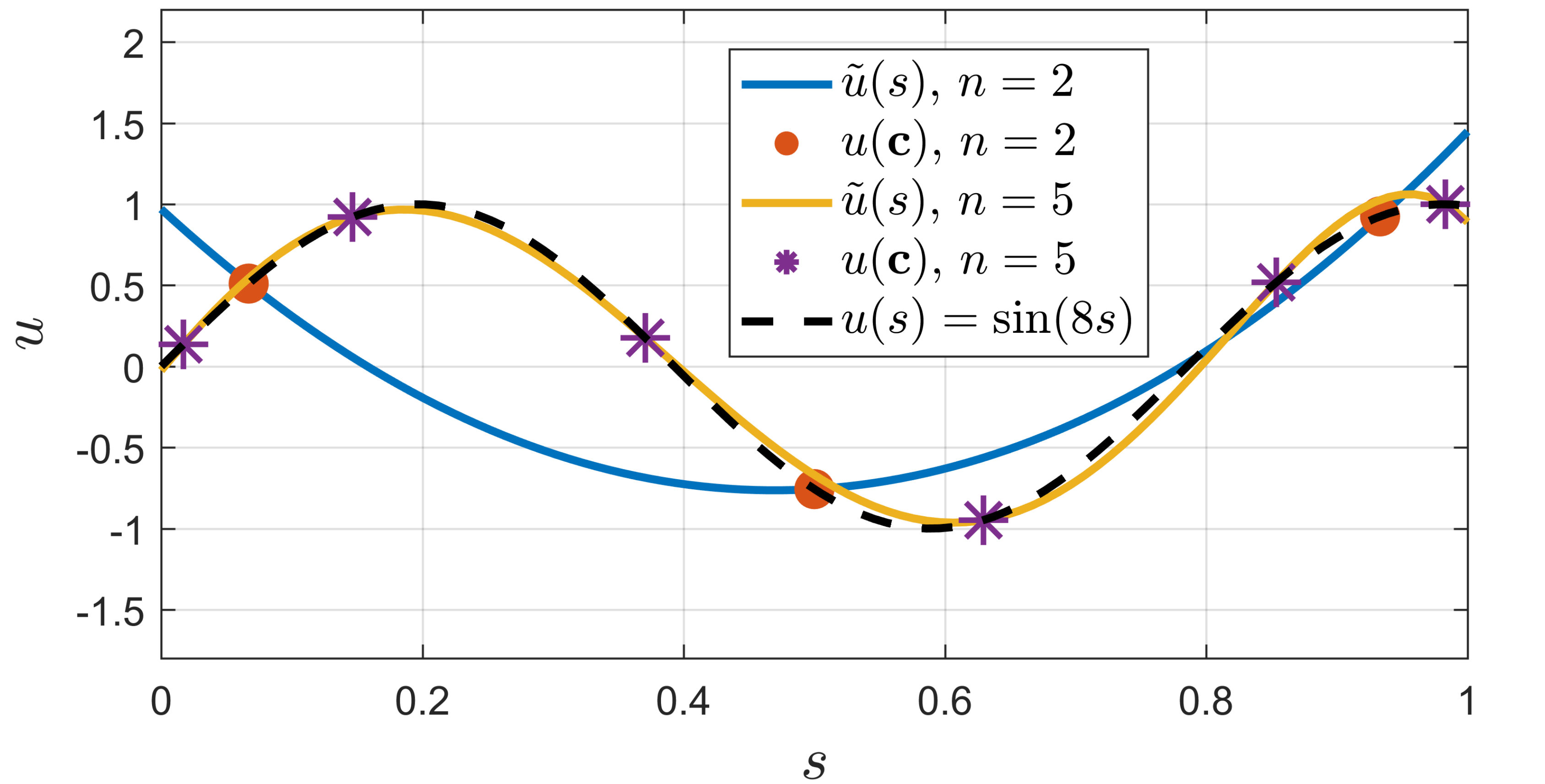}
\caption{Example of polynomial interpolation of a function $u(s) = \sin(8s)$ using a 2nd order and 5th order Chebyshev polynomials with interpolation nodes at the zeros of the $n$th order Chebyshev polynomial. Orthogonal interpolation provides rapid convergence with increasing $n$.}
\label{fig:collocation_example}
\end{figure}
\par Any continuous function can be approximated by obtaining the function values at a set of \emph{interpolation nodes}, then fitting an \emph{interpolating polynomial} to those function values. If we choose an interpolating polynomial of sufficiently high order, it can provide a reasonable approximation of the function. Figure \ref{fig:collocation_example} shows an example of this, where $u(s) = \sin(8s)$ is approximated to varying degrees of fidelity by $\tilde{u}(s)$ given as a 2nd order polynomial or a 5th order polynomial. To avoid Runge's phenomenon, the interpolation nodes are chosen to be the zeros of an orthogonal polynomial. Here we choose Chebyshev polynomials but other orthogonal polynomials could also be used.
\par In a collocation method, an interpolating polynomial is used to find an approximate solution to a DE. An interpolating polynomial is chosen to approximate the unknown solution to the DE (which in our case is $\bs{\xi}(s)$, the unknown twist distribution), a set of \emph{collocation points} (similar to the interpolation nodes above) are chosen, and it is enforced that the interpolating polynomial satisfy the boundary conditions as well as the DE at the collocation points. This results in a set of algebraic equations that can be solved using standard nonlinear root-finding approaches. We now discuss how to carry out this procedure for a Cosserat rod.
\par  In our context, we seek to find the unknown twist distribution $\bs{\xi}(s)$ that satisfies the ODEs (\ref{eq:cosserat_ode}) and the boundary conditions (\ref{eq:boundary_condition}). We choose to describe this unknown twist distribution as a set of $n$th order Chebyshev polynomials of the first kind, denoted as $\tilde{\bs{\xi}}(s) = [\tilde{\mb{u}}(s),0,0,1]\T$, where $\tilde{\mb{u}}(s) = [\tilde{u}_x(s),\tilde{u}_y(s),\tilde{u}_z(s)]\T, s\in[0,L]$. Note that here we use three interpolating polynomials to approximate $\bs{\xi}(s)$, but for a general Cosserat rod case where shear strains are included, six interpolating polynomials would be needed. We choose the collocation points $\mb{c} = [c_0,\dots,c_n]\T$, $c_i\in[0,L], \, i=0\ldots n$ to be the Chebyshev polynomial zeros and call the function values $\tilde{\mb{u}}(c_i)$ the \emph{collocation values}.
\par The Chebyshev polynomials of the first kind can be conveniently represented by a recurrence relation \cite{mason2002chebyshev}:
\begin{equation} \label{eq:cheb_def}
\begin{gathered}
T_n(x) = 2x\,T_{n-1}(x) - T_{n-2}(x), \quad n=2,3,\dots \\
T_0 = 1, \quad T_1(x) = x
\end{gathered}
\end{equation}
where $T_n(x), x \in [-1,1]$ is the $n$th Chebyshev polynomial. To shift the Chebyshev polynomials to the domain $s \in [0,L]$ we apply the linear transformation
\begin{equation} \label{eq:shifted_cheb}
x(s) = \frac{2s - L}{L}
\end{equation}
and evaluate $T_n(x(s))$ via (\ref{eq:cheb_def}). Henceforth, we denote $T_n(x(s))$ as simply $T_n(s)$ with (\ref{eq:shifted_cheb}) implied.
\par In a typical direct variational method, one describes the interpolating polynomial as an expansion in a basis of orthogonal polynomials, which requires computing modal coefficients from collocation values. For example, $\tilde{u}_{x}(s)$ may be represented by the $n$th order interpolating polynomial:
\begin{equation} \label{eq:interp_poly}
\tilde{u}_{x}(s) = \frac{1}{2}a_0T_0(s) + \sum_{i = 1}^{n} a_iT_i(s)
\end{equation}
where the modal coefficients $a_i$ are found via:
\begin{equation} \label{eq:cheb_coeff}
a_i = \frac{2}{n+1}\sum_{k = 0}^{n}\tilde{u}_x(c_k)T_i(c_k)
\end{equation}
In an orthogonal collocation method, computing the modal coefficients is avoided by using a \emph{differentiation matrix}. By taking advantage of the discrete orthogonality of Chebyshev polynomials, it has been shown that the derivatives at the collocation points can be written as a linear combination of the collocation values \cite{mason2002chebyshev,villadsen1978approximation}:
\begin{equation} \label{eq:diff_mat}
\frac{d}{ds}\begin{pmatrix} u_x(c_0) \\ \vdots \\ u_x(c_n)  \end{pmatrix} = \mb{D}_n\begin{pmatrix} u_x(c_0) \\ \vdots \\ u_x(c_n)  \end{pmatrix}
\end{equation}
where the element in row $(i+1)$ and column $(j+1)$ of $\mb{D}_n$ is given by:
\begin{equation}
d_{ij} = \left\{\begin{array}{lr}
\frac{1}{2}\frac{T''_{n+1}(c_i)}{T'_{n+1}(c_i)}, & i = j \\[8pt]
\frac{T'_{n+1}(c_i)}{(c_i-c_j)T'_{n+1}(c_j)}, & i \neq j \\[8pt]
\end{array}\right.
\end{equation}
Note that the above result requires that the collocation points be the Chebyshev zeros and that $\mb{D}_n$ can be computed offline if the order of the interpolating polynomial is chosen \emph{a priori}.
\par We now assemble the collocation values into a matrix $\mb{U}_c$ where each column contains an interpolating polynomial:
\begin{equation}
\mb{U}_c = \begin{pmatrix}
\tilde{u}_x(c_0)&\tilde{u}_y(c_0)&\tilde{u}_z(c_0)\\
\vdots & \vdots & \vdots \\
\tilde{u}_x(c_n)&\tilde{u}_y(c_n)&\tilde{u}_z(c_n)\\
\end{pmatrix} = \begin{pmatrix}
\tilde{\mb{u}}\T(c_0) \\
\vdots \\
\tilde{\mb{u}}\T(c_{n})
\end{pmatrix}
\end{equation}
We want the interpolating polynomials to satisfy the boundary conditions and (\ref{eq:cosserat_ode}) at the collocation points. To ensure a square error residual Jacobian, we remove the last row of $\mb{D}_n$ to form $\mb{D}_{n-1}$ and form a matrix of error residuals:
\begin{equation} \label{eq:full_coll_system}
\mb{E} = \begin{pmatrix} \mb{D}_{n-1}\mb{U}_c \\
\tilde{\mb{u}}\T(L)
\end{pmatrix}
- \begin{pmatrix} \mb{g}\T(\tilde{\mb{u}}(c_0)) \\ \vdots \\ \mb{g}\T(\tilde{\mb{u}}(c_{n-1})) \\ \left(\mb{K}^{-1}\mb{R}\T(L)\mb{m}_e\right)\T\end{pmatrix} \end{equation}
where we evaluate $\mb{g}(\tilde{\mb{u}}(c_i))$ by plugging the collocation value $\tilde{\mb{u}}(c_i)$ into (\ref{eq:cosserat_ode}). Note that $\mb{u}(L)$ is not one of the collocation values and must be computed from the interpolating polynomial. We show how to do this in Section \ref{sec:interpolation}.
\par Equation (\ref{eq:full_coll_system}) is a set of nonlinear algebraic equations we must now solve for the collocation values $\tilde{\mb{u}}(c_i), i \in 0,\dots,n$. This is achieved by minimizing the error in (\ref{eq:full_coll_system}) using a nonlinear solver (e.g. Levenberg-Marquardt). Specifically, the error is defined as $\mb{e} = \text{vec}(\mb{E})$, where $\text{vec}(\mb{E})$ arranges the columns of $\mb{E}$ into a vector. Once $\tilde{\mb{u}}(c_i)$ are found, the modal coefficients and the interpolating polynomial $\tilde{\mb{u}}$ are defined. Therefore, $\mb{X}(s)$ is defined in terms of the Chebyshev polynomials.
\par Computing $\mb{g}(\tilde{\mb{u}}(c_i))$ in (\ref{eq:full_coll_system}) requires integrating $\mb{T}'(s)$ to find $\mb{R}(s)$ at the collocation points. Examples of methods to do this include explicit Runge-Kutta methods, quaternion integration \cite{rucker2018integrating}, and a variety of geometric integration methods \cite{park2005geometric,iserles2000lie}. Although any of these approaches could be combined with collocation, here we take the geometric integration approach. We observe that $\mb{T}(s)$ is an element of the Lie group $SE(3)$, $\mb{X}(s)$ is an element of the Lie algebra $se(3)$, and that given a set of collocation values $\tilde{\mb{u}}(c_i)$ (which would be guessed at each iteration of a nonlinear solver), the twist distrubtion $\mb{X}(s)$ is known via the interpolating polynomial. Therefore, $\mb{T}'(s)$ in (\ref{eq:cosserat_ode}) represents a linear DE on a Lie group, a class of problems that has received extensive study in other work \cite{iserles2000lie}. We use a known result from the Lie group integration literature called the \textit{Magnus expansion} to forward integrate $\mb{T}'(s)$. Note that a similar Lie algebra expansion was given in \cite{muller_approximation_2010}, which could also be used here in place of the Magnus expansion.
\par The benefits of using the Magnus expansion are 1) the integration is done on the Lie algebra $se(3)$ and mapped to $SE(3)$ via an exponential mapping, so $\mb{T}(s)$ is guaranteed to stay on $SE(3)$, 2) the quadrature method given in \cite{iserles2000lie} for the Magnus expansion is numerically efficient and allows for closed-form gradients to be computed, and 3) the resulting solution to the BVP is given as a product of matrix exponentials that can be used for further analytical study (e.g. computing Jacobians). In the next section, we provide the main results on the Magnus expansion without proof and refer the reader to \cite{iserles2000lie} for additional details.
%===============================================================================
\section{The Magnus Expansion} \label{sec:magnus}
It was shown in \cite {iserles2000lie} that for sufficiently small $s$ the solution to $\mb{T}'(s) = \mb{X}(s)\mb{T}(s)$ can be expressed as a matrix exponential of a twist $\bs{\psi} = [\bs{\psi}_u,\bs{\psi}_v]\T \in \realfield{6}$:
\begin{equation}
\begin{aligned}
\mb{T}(s) &= \mb{T}_0e^{\mb{\Psi}(s)} \\
\mb{\Psi}(s) \triangleq \widehat{\bs{\psi}}(s) &= \begin{pmatrix}
\widehat{\bs{\psi}}_u(s) & \bs{\psi}_v(s) \\
0 & 0
\end{pmatrix} \in se(3)
\end{aligned}
\end{equation}
A short proof in \cite{iserles2000lie} shows that the twist matrix $\mb{\Psi}(s)$ satisfies the differential equation
\begin{equation} \label{eq:psi_dot}
\mb{\Psi}'(s) = \text{dexp}^{-1}_{-\mb{\Psi}(s)}(\mb{X}(s)) \quad \mb{\Psi}(0) = \mb{0}
\end{equation}
where the $\text{dexp}$ operator is defined as:
\begin{equation} \label{eq:dexp}
\text{dexp}^{-1}_{\mb{\Psi}} = \sum_{i = 0}^{\infty} \frac{B_i}{i!}\text{ad}^i_{\mb{\Psi}}
\end{equation}
where $B_i$ are the Bernoulli numbers and $\text{ad}_{\mb{\Psi}}$ denotes the $6\times6$ adjoint representation of an element in $se(3)$ \cite{selig2004geometric}:
\begin{equation}
\text{ad}_{\mb{\Psi}} = \begin{pmatrix}
\widehat{\bs{\psi}}_u(s) & 0 \\
\widehat{\bs{\psi}}_v(s) & \widehat{\bs{\psi}}_u(s)
\end{pmatrix}
\end{equation}
To compute $\text{dexp}^{-1}_{-\mb{\Psi}}$, one can either truncate the infinite series or, for special cases, derive a closed-form expression as done in \cite{selig2004geometric} for $se(3)$. The DE in (\ref{eq:psi_dot}) can then be numerically integrated with a standard Runge-Kutta method to find $\mb{\Psi}(s)$. To reduce the cost of numerically integrating, we take an alternative approach that avoids computing (\ref{eq:dexp}). In \cite{magnus1954exponential}, Magnus solved (\ref{eq:psi_dot}) via Picard iteration, leading to a solution for $\mb{\Psi}(s)$ written as an infinite series of terms consisting of integrals of commutators. In \cite{iserles2000lie}, order analysis showed which terms can be dropped for a given order, resulting in the following fourth order Magnus expansion:
\begin{equation} \label{eq:mag_fourth}
\begin{aligned}
\bs{\Psi}^{[4]}(s) = &\int_{0}^{s}\mb{X}(\eta) \, \text{d}\eta  \\
&+ \frac{1}{2} \int_{0}^{s} \left[\int_0^{\eta_1} \mb{X}(\eta_2) \; \text{d}\eta_2,\mb{X}(\eta_1)\right]\text{d}\eta_1
\end{aligned}
\end{equation}
where the matrix commutator is given by $\left[ \mb{X}_1, \mb{X}_2 \right] = \mb{X}_1\mb{X}_2 - \mb{X}_2\mb{X}_1$.
We do not replicate it here, but \cite{iserles2000lie} also provides the sixth order expansion with 7 terms and up to 4 integrals per term.
\par At first glance, the fourth and sixth order Magnus expansions seem expensive to compute, however, it was shown in \cite{iserles2000lie} that both expansions can be efficiently computed with Gaussian quadrature. Gaussian quadrature is an approach for approximating definite integrals via interpolating polynomials and leads to expressing the integral as a weighted sum of the function values, which we call \emph{quadrature values}, evaluated at the \emph{quadrature points}, which are chosen to be the zeros of an orthogonal polynomial.
\par In addition to showing how to compute the expansion via Gaussian quadrature, it was shown in \cite{iserles2000lie} that skew symmetry of the commutators allows many of the terms in the quadrature to be combined. It was also shown via order analysis that if the quadrature points are chosen to be symmetric about $\frac{1}{2}h$, where $h$ is width of interval between two adjacent collocation points, many of the terms in the quadrature can be dropped for a given order. We first provide here the steps to compute the quadrature on the interval $[0,h]$ (as given in \cite{iserles2000lie}) then describe how this can be combined with the collocation method above.
\par To integrate between collocation point $c_i$ and $c_{i+1}$, we choose the quadrature interpolating polynomials to be the Legendre polynomials shifted to $[0,1]$ and choose $\nu$ quadrature points $\mb{t} = [t_1,\dots,t_\nu]\T \in [0,1]$ to be the zeros of the Legendre polynomials since they are symmetric about $\frac{1}{2}$. We then form the quadrature values:
\begin{equation}
\mb{X}_k = h\mb{X}(c_i+t_k h), \quad k = 1,2,\dots,\nu
\end{equation}
Note that an order four quadrature requires $\nu \geq 2$ and an order six quadrature requires $\nu \geq 3$. In \cite{iserles2000lie,munthe1999computations}, a change of basis is carried out to take advantage of the symmetry of the Magnus expansion. The change of basis is done by finding the solution of the following Vandermonde system:
\begin{equation}
\sum_{i = 1}^{\nu}\left(t_k - \tfrac{1}{2} \right)^{i-1}\,\mb{Y}_i = \mb{X}_k
\end{equation}
where $\mb{Y}_i \in se(3)$ are solved for by inversion:
\begin{equation}
\mb{V}_{ij} = \left(t_i - \tfrac{1}{2} \right)^{j-1}, \quad\, \mb{Y}_i = \sum_{j = 1}^{\nu}\left(\mb{V}^{-1}\right)_{ij}\mb{X}_j
\end{equation}
This leads to a quadrature rule for an order four expansion:
\begin{equation}\label{eq:quadrature_4th}
\mb{\Psi}^{[4]}(h) = \mb{Y}_1 + \frac{1}{12}[\mb{Y}_1,\mb{Y}_2]
\end{equation}
and the following quadrature rule for a sixth order expansion:
\begin{multline}\label{eq:quadrature_6th}
\mb{\Psi}^{[6]} = \mb{Y}_1 + \frac{1}{12}\mb{Y}_3 + \frac{1}{12}[\mb{Y}_1,\mb{Y}_2] \\ - \frac{1}{240}[\mb{Y}_2,\mb{Y}_3 ] + \frac{1}{360}[\mb{Y}_1,[\mb{Y}_1,\mb{Y}_3] \\ -\frac{1}{240}[\mb{Y}_2,[\mb{Y}_1,\mb{Y}_2]] - \frac{1}{720}[\mb{Y}_1,[\mb{Y}_1,[\mb{Y}_1,\mb{Y}_2]]]
\end{multline}
Note that in \cite{iserles2000lie} the expansion is given for the form of $\mb{T}'(s) = \mb{X}(s)\mb{T}(s)$, which corresponds to twists expressed with respect to the world frame, but since we use the body twist in (\ref{eq:cosserat_ode}) there is a difference in sign for some terms.
\par
\begin{figure}[tbp]
\center
\includegraphics[width=0.9\columnwidth]{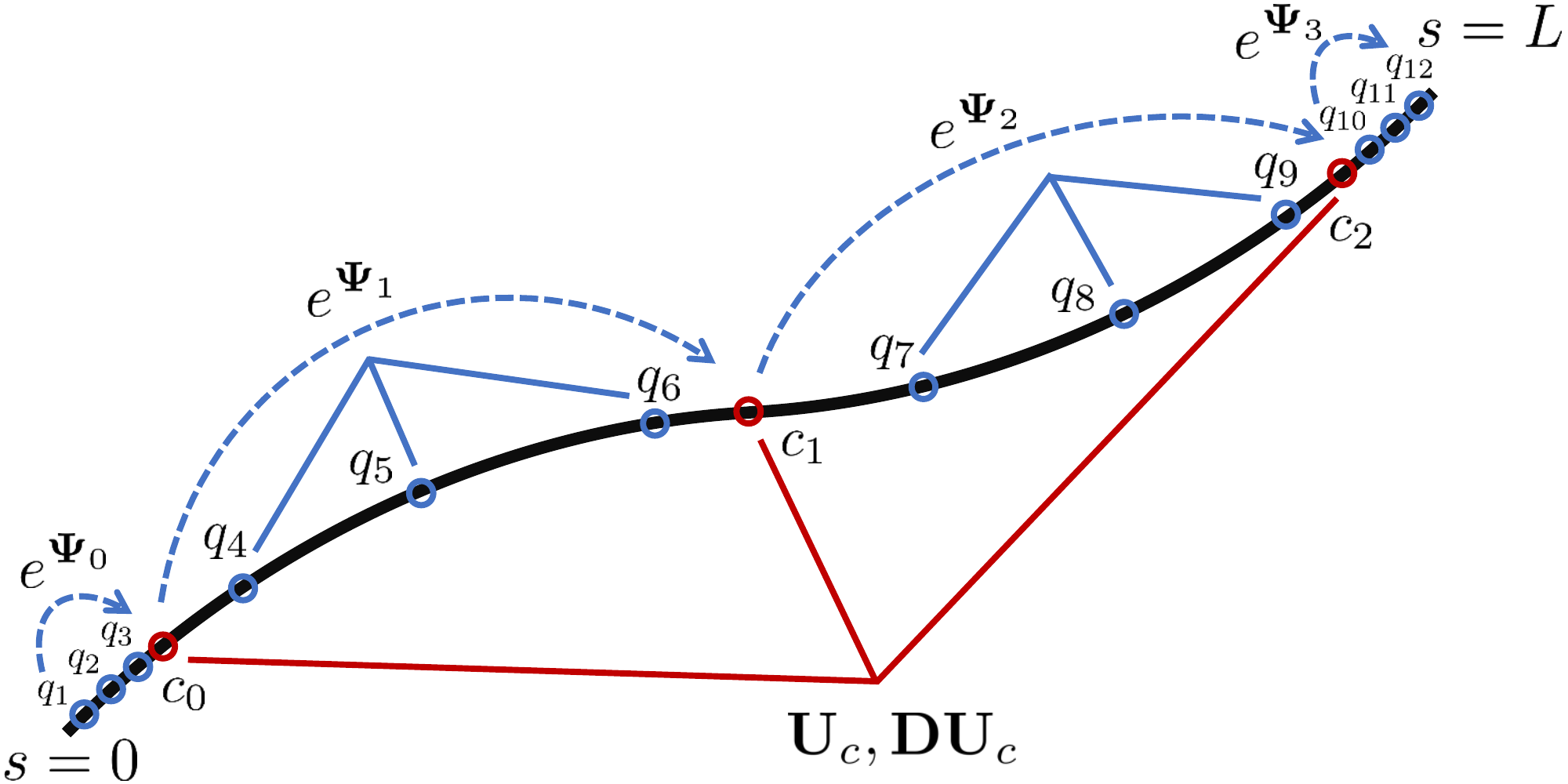}
\caption{Collocation points and quadrature points  for the case of $n = 2$ and $\nu = 3$. Using orthogonal polynomials and their zeros results in quadrature values being linearly related to the collocation values.}
\label{fig:coll_quad_pt}
\end{figure}

 We propose finding the poses at the collocation points $\mb{T}(c_i)$ using a Magnus expansion step between each collocation point. We assign either $\nu = 2$ or $\nu = 3$  quadrature points (for either fourth or sixth order expansions, respectively) between each pair of collocation points as well as between $0$ and $c_0$ and between $c_n$ and $L$. This leads to a total of $m = \nu(n+2)$ quadrature points along the length of the rod. Fig. \ref{fig:coll_quad_pt} shows an example of this for $\nu = 3$ and $n=2$. We then compute a Magnus expansion between each collocation point, starting from $s=0$ and stepping to $s=L$. The frame at each collocation point therefore given by a product of exponentials where $\mb{\Psi}_i$ is expressed in frame $\mb{T}(c_{i})$:

\begin{equation} \label{eq:prod_exp}
\mb{T}(c_k) = \mb{T}_0 \prod_{i = 0}^{k} e^{\mb{\Psi}_i}
\end{equation}
Although not detailed here, we note that since the shape of the rod is given as a product of exponentials, closed-form gradients of the residual vector $\mb{e}$ can be found which facilitate faster solutions to (\ref{eq:full_coll_system}). Furthermore, as shown in \cite{renda_discrete_2018,grazioso_geometrically_2018} Jacobians useful for studying the kinematics of continuum manipulators can be found via (\ref{eq:prod_exp}).
\section{Determining curvature at quadrature points} \label{sec:interpolation}
Evaluating $\mb{\Psi}_i$ according to the quadrature formula above ((\ref{eq:quadrature_4th}) or (\ref{eq:quadrature_6th})) requires knowing the values at the quadrature points, $\tilde{\mb{u}}(q_k)$ where $q_k=c_i+t_k\,h$, $k=1\ldots \nu$. Given the collocation values $\tilde{\mb{u}}(c_i)$ (which are guessed at each iteration of a nonlinear solver), we show here that the function values of the interpolating polynomial at the quadrature points can be found directly as a linear combination of the collocation values. First, we define a matrix $\mb{U}_q$ that contains quadrature values, similar to $\mb{U}_c$:
\begin{equation}
\mb{U}_q = \begin{pmatrix}
\mb{u}\T(q_1) \\
\vdots \\
\mb{u}\T(q_{m})
\end{pmatrix}
\end{equation}
We then use (\ref{eq:interp_poly}) together with (\ref{eq:cheb_coeff}) to find a matrix relating the collocation values and the quadrature values:
\begin{equation}
\mb{U}_q =  \mb{A}\mb{B}\mb{U}_c
\end{equation}
where the matrix $\mb{B} \in \realfield{(n+1)\times(n+1)}$ transforms the collocation values into modal coefficients which are then a transformed by $\mb{A} \in \realfield{m\times(n+1)}$ into quadrature values:
\begin{equation}
\mb{A} = \begin{pmatrix}
\frac{1}{2}T_0(q_1) & T_1(q_1) & \dots & T_n(q_1) \\
\vdots & \vdots & \vdots & \vdots\\
\frac{1}{2}T_0(q_m) &  T_1(q_m) & \dots & T_n(q_m) \\
\end{pmatrix}
\end{equation}
\begin{equation}
\mb{B} = \frac{2}{n+1}\begin{pmatrix}
T_0(c_0) & \dots & T_0(c_{n}) \\
\vdots & \vdots & \vdots \\
T_n(c_0) & \dots & T_n(c_{n}) \\
\end{pmatrix}
\end{equation}
This same procedure is also used to find $\mb{u}(L)$ which is necessary to compute the error residual for the boundary condition in (\ref{eq:full_coll_system}). Note that both $\mb{A}$ and $\mb{B}$ can be computed offline as long as the collocation and quadrature points are chosen beforehand.
\begin{figure*}[t]
\center
\includegraphics[width=0.92\textwidth]{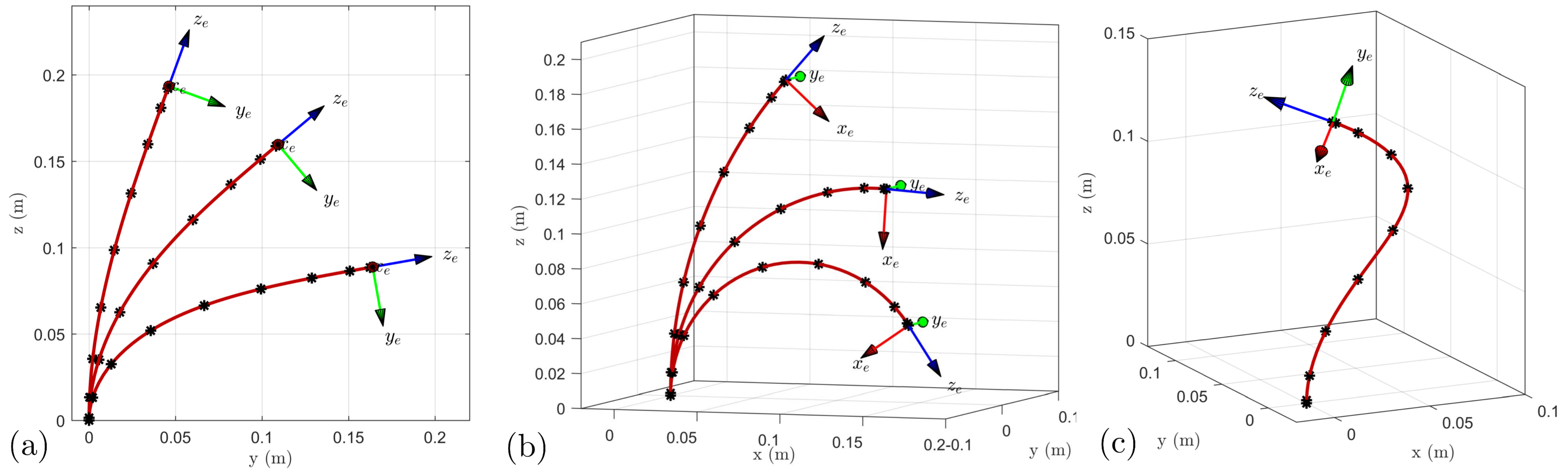}
\caption{Examples comparing the shooting method and our collocation approach for (a) force loads (which we also validate with elliptic integrals) (b) moment loads which result in constant-curvature, and (c) combined force and moment.}
\label{fig:simple_examples}
\end{figure*}
\section{Convergence of the Magnus Expansion}
One important note is that the infinite Magnus expansion is not guaranteed to converge. In this section, we use known sufficient conditions for convergence of the Magnus expansion to provide the maximum step sizes that will guarantee convergence. These bounds are not necessary conditions, so we are able to violate the bounds in our simulation results, but we discuss convergence here since it cannot in general be guaranteed. For real matrices, the expansion converges in the 2-norm provided that \cite{moan2008convergence}:
\begin{equation}
\int_0^{h}\| \mb{X}(\xi) \|_2 \; \text{d}\xi < \pi
\end{equation}
Here we relate this bound on convergence to the case of the Cosserat rod and provide a bound on the maximum step size as a function of the maximum curvature. Recall that the Euclidian 2-norm is bounded by the Frobenius norm:
\begin{equation}
\| \mb{X} \|_2 \leq \| \mb{X} \|_F = \sqrt{\text{trace}(\mb{X}\T\mb{X})}
\end{equation}
Assume the curvatures $\mb{u}(s)$ are bounded by a known scalar, i.e. $u_x \leq u_y \leq u_z \leq \beta$. We can then compute the Frobenius norm explicitly and provide the following bound:
\begin{equation} \label{eq:frobenius_bound}
\| \mb{X} \|_F = \sqrt{2u_x^2 + 2u_y^2 + 2u_z^2 + 1} \leq \sqrt{6\beta^2 + 1}
\end{equation}
We then integrate (\ref{eq:frobenius_bound}) to find a conservative bound for integration step $h$ to guarantee convergence:
\begin{equation}
\int_0^{h}\| \mb{X}(\xi) \|_2 \; \text{d}\xi \leq \int_0^{h}\| \mb{X}(\xi) \|_F \; \text{d}\xi < \pi
\end{equation}
\begin{equation}
h_{max} < \frac{\pi}{\sqrt{6\beta^2 + 1}}
\end{equation}
\par Given a particular task and continuum robot architecture, the bound $\beta$ might be known beforehand through a simulation study. Another option is to choose $\beta$ by considering the strain limits of the rod material. Consider as an example a superelastic nickel-titanium (Nitinol) rod, which can accept a strain of 5\% with minimal loss of superelasticity due to cyclic fatigue \cite{NiTi_fatigue2006}. The maximum bending strain is $\epsilon = ur$, where $u$ is the rod's curvature and $r$ is the rod's radius. Assuming the continuum robot is designed to avoid violating strain limits, we have $u_x \leq u_y \leq u_z \leq \epsilon / r = \beta$.

\begin{table} [h]
\caption{Maximum Step Size for Gauranteed Convergence  with 6\% Bending Strain}
\centering \label{table:h_max}
\begin{tabular}{|c|c|c|} \hline
$r$ (mm) & $\beta$ & $h_{max}$ (mm) \\ \hline
1 & 50 & 25.65 \\ \hline
2 & 25 & 51.29 \\ \hline
3 & 16.67 & 76.92 \\ \hline
4 & 12.5 & 102.54\\ \hline
\end{tabular}
\end{table}
\begin{table} [ht]\caption{Step Sizes Used in Simulations (2 mm OD rod)} \label{table:step_sizes}
\centering \begin{tabular} {|c|c|c|c|c|c|c|} \hline
$n$ & 2 & 4 & 6 & 8 & 10 \\ \hline
$h_{max}$ (mm) & 86.60 & 58.78 & 43.38 & 34.20 & 28.17 \\ \hline
\end{tabular}
\end{table}

\par Table \ref{table:h_max} shows the maximum step size that will guarantee convergence of the Magnus expansion under the assumption $\epsilon < 5$\%. For larger rod diameters the step sizes are not restrictive, and for smaller rod diameters the particular task can potentially be taken into account to determine a more suitable bound $\beta$ and allow larger step sizes. Online checks may be necessary to check for convergence of the expansion in cases where larger step sizes are used. In our simulations, we used the step sizes in Table \ref{table:step_sizes} without observing issues.

%===============================================================================
\section{Simulation Results} \label{sec:results}
In this section, we will compare the accuracy of the fourth and sixth-order Magnus expansions with different numbers of collocation points. In all examples we use a shooting method as a ground-truth to compare our results against, since the shooting method has been validated in previous experimental studies \cite{chikhaoui_comparison_2019,till_efficient_2015,orekhov_modeling_2017}. We integrated (\ref{eq:cosserat_ode}) with a Runge-Kutta solver and a Levenberg-Marquardt algorithm (\emph{ode45}() and \emph{fsolve}() in MATLAB) to satisfy the boundary conditions with a termination tolerance of $10^{-9}$. We used units of meters and radians. In the planar examples with in-plane forces, we also compared against the elliptic integral solution given in \cite{howell1995parametric} and experimentally validated in \cite{xu2010analytic,wei_modeling_2012}. For all examples we simulated the case of a solid Nickel-titanium (Nitinol) rod with a 2 mm diameter and a 200 mm length.
\par We first consider the planar case with an in-plane load. We solved the elliptic integral equations in \cite{howell1995parametric} with tip angles of $20$\textdegree, $50$\textdegree, and $80$\textdegree, giving the position of the tip and resultant tip forces of $\mb{f}_e = [0,1.04,0.104]\T$ N, $\mb{f}_e = [0, 3.63,0.362]\T$ N, and $\mb{f}_e = [0,18.9,1.89]\T$ N. We then used these forces in the shooting method and our collocation method (with a sixth order Magnus expansion and $n = 10$), both of which are shown in Fig. \ref{fig:simple_examples}(a). All three methods showed agreement within $0.006$ mm (0.003\% of arc-length).
\par Next we consider a planar case with an in-plane moment load, as shown in Fig. \ref{fig:simple_examples}(b). In this case, our collocation method returns the predicted constant-curvature shape, and for all three cases shown in Fig. \ref{fig:simple_examples}(b) the shooting and collocation method (with a sixth order Magnus expansion and $n = 10$) showed agreement within 3.98e-6 mm and rotation error within machine precision.
\begin{figure}[htbp]
\center
\includegraphics[width=0.84\columnwidth]{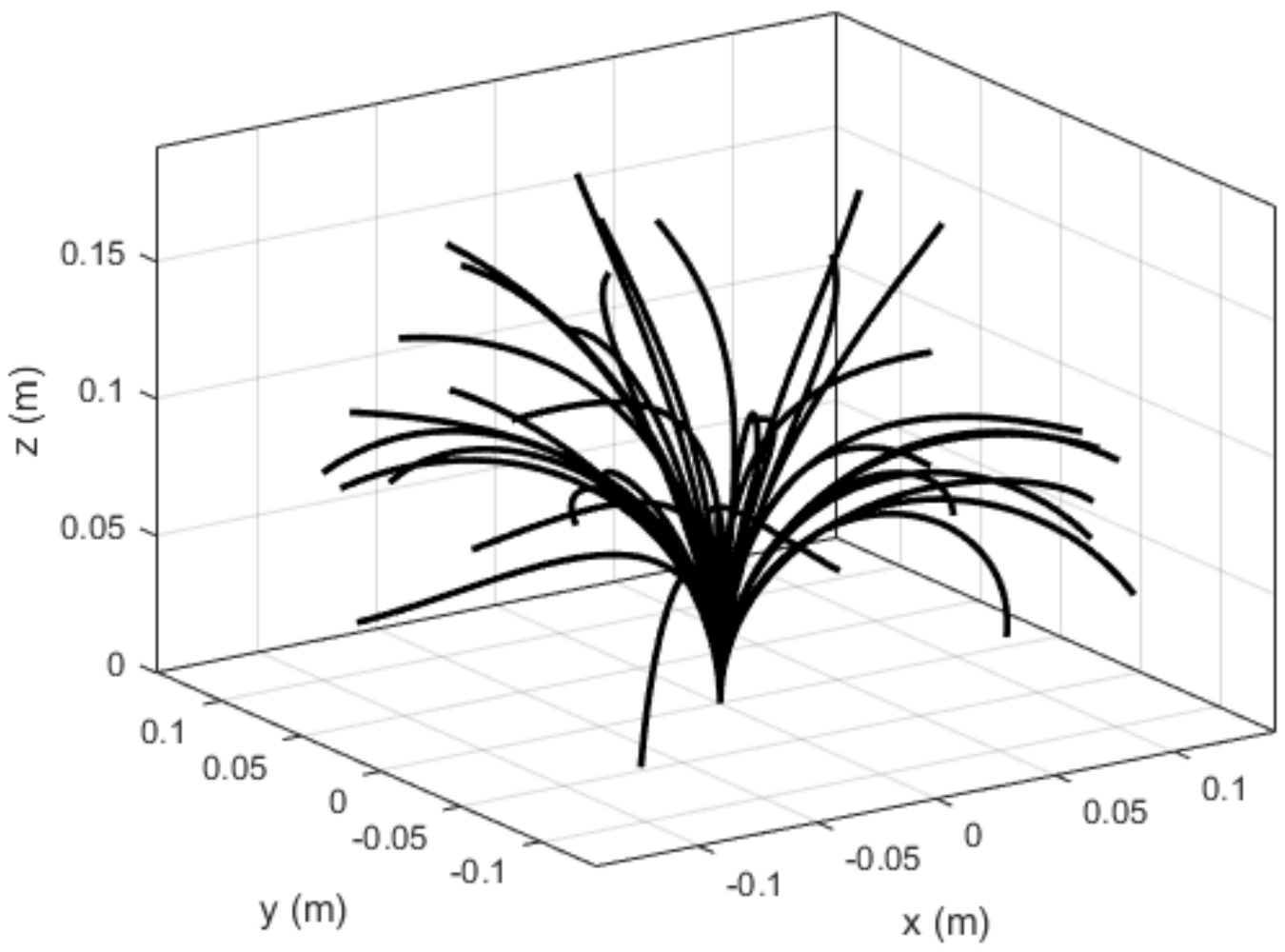}
\caption{Samples from the set of 729 rod shapes used when comparing our approach to the shooting method.}
\label{fig:workspace_samples}
\end{figure}
\par The collocation method also agrees with the shooting method in cases with combined forces and moments, as shown in Fig. \ref{fig:simple_examples}(c). In the next study, we compare the shooting and collocation methods for combinations of forces/moments and compare different numbers of collocation points. We simulated combinations of $\pm1$ N tip forces and $\pm 0.5$ Nm moments applied to the tip of the rod (a total of 729 applied wrench cases). The solution speed of any BVP solver is affected by how close initial guesses are to the solution, so for each case, we begin the simulation with the rod in its straight configuration and linearly interpolate from a zero wrench to the final wrench in 3 steps. This is to more realistically simulate a scenario where the model is being continuously solved (e.g. kinematic control) and the previous solution is used as the initial guess for the solver. With the interpolated wrenches the total number of applied wrench cases is 2,187, samples of which are shown in Fig. \ref{fig:workspace_samples}. We solved for the equilibrium shape of the rod for each case with different numbers of collocation points, $n \in [2, 4, 6, 8, 10]$. We also solved each case using the fourth-order Magnus expansion ($\nu = 2$) and the sixth-order Magnus expansion ($\nu = 3$). The tip pose found with each collocation method was compared to the tip pose found with the shooting method using the following metrics:
\begin{equation}
\begin{aligned}
e_p &= \frac{\| \mb{p}_c(L) - \mb{p}_s(L)\|}{L} \times 100 \\
e_r &= \cos^{-1}\left( \frac{\text{trace}\left(\mb{R}_s(L)\mb{R}\T_c(L)\right)-1}{2}  \right)
\end{aligned}
\end{equation}
where we report the position error as percent of total arc-length, $\mb{p}_c(L),\mb{R}_c(L)$ are the tip position/rotation found using a collocation method, and $\mb{p}_s(L),\mb{R}_s(L)$ are the tip poses found using the shooting method.
\par Tables \ref{table:errors_fourth} and \ref{table:errors_sixth} show the results of the simulations. The difference between the shooting method and our method rapidly converges to zero with increasing $n$. Both the fourth-order and sixth-order expansions provided rapid convergence with increasing $n$, and for $n\leq6$ are comparable. However, for $n\geq8$ sixth-order Magnus steps did a better job reducing the $e_p$ and $e_r$ to below 0.005\%. This implies that for small $n$, the primary source of the error is in the interpolation error in collocation (and not the Magnus stepping), while for larger $n$ the error in the Magnus steps becomes more important. Any method with $n\geq6$ was able to calculate the tip position with agreement with shooting of less than 0.15\% of arc length.
\par We also report in Tables \ref{table:errors_fourth} and \ref{table:errors_sixth} the average speed at which solutions were found across the set of 2,187 shapes, running MATLAB 2019b on an i7-4770 3.4 GHz CPU. The fourth order and sixth order Magnus expansions were comparable in terms of speed, since the primary source of computation effort is not in evaluating the Magnus expansion but in computing the gradient of the error residual in (\ref{eq:full_coll_system}), which despite being available in closed-form, becomes increasingly expensive as $n$ increases. When $n$ is not too large, the gradient is cheap to compute which provides quick solutions when combined with the efficient integration of $\mb{T}'(s)$ via the Magnus expansion.
\begin{table} [h]\caption{Fourth Order Magnus Tip Error as a Function of Collocation Polynomial Order ($L = 200$ mm)} \label{table:errors_fourth}
\centering \begin{tabular} {|c|C{0.9cm}|C{0.9cm}|C{0.9cm}|C{0.9cm}|C{1.3cm}|} \hline
 & \multicolumn{2}{c|}{Pos. $e_p$ (\%)} & \multicolumn{2}{c|}{Rot. $e_r$ (deg)} & \\ \hline
& Avg. & Max & Avg. & Max. & Speed (Hz) \\ \hline
$n = 2$  & 2.97 &  28.0 & 4.28  & 36.3 & 179.6\\ \hline
$n = 4$  & 0.141 & 2.15 & 0.235  & 3.78 & 112.1\\ \hline
$n = 6$  & 0.00573  & 0.147  & 0.00889  & 0.183 & 71.6\\ \hline
$n = 8$  & 0.00122 & 0.0173 & 0.00453 & 0.0571 & 46.3\\ \hline
$n = 10$ & 5.46e-4& 0.00707 & 0.00448 & 0.0543 &  33.1\\ \hline
\end{tabular}
\end{table}
\begin{table} [h]\caption{Sixth Order Magnus Tip Error as a Function of Collocation Polynomial Order ($L = 200$ mm)} \label{table:errors_sixth}
\centering \begin{tabular} {|c|C{0.9cm}|C{0.9cm}|C{0.9cm}|C{0.9cm}|C{1.3cm}|} \hline
 & \multicolumn{2}{c|}{Pos. $e_p$ (\%)} & \multicolumn{2}{c|}{Rot. $e_r$ (deg)} & \\ \hline
& Avg. & Max. & Avg. & Max. & Speed (Hz)\\ \hline
$n = 2$& 3.00  & 28.1 & 4.29 & 36.5 & 176.8\\ \hline
$n = 4$&  0.140  & 2.26 & 0.234 & 3.79 & 106.2\\ \hline
$n = 6$&  0.00467 & 0.115 & 0.00889 & 0.193 & 68.8\\ \hline
$n = 8$&  1.95e-4 & 0.00493 & 0.00450 & 0.0553 &  42.5\\ \hline
$n = 10$& 2.66e-5 & 0.00140 & 0.00448 & 0.0542 & 32.4\\ \hline
\end{tabular}
\end{table}
\par Our shooting method implementation in MATLAB solved the BVPs at an average rate of 17.6 Hz (with residual gradients estimated via finite differences). Our collocation implementation was faster than the shooting method in all examples, but we would like to stress that a fair comparison between the shooting and collocation is difficult since either method could be improved with more specialized implementations in pre-compiled code and by providing good initial guesses (see \cite{till_efficient_2015,dupont2009design} for shooting method implementations with >1kHz). The main takeaway from these results is that our method can be competitive in terms of computation speed, and due to the availability of closed-form expressions, may improve speed in cases where lower-order polynomials are sufficient. More importantly, the Magnus expansion leading to closed-form direct kinematics offers fast evaluation of instantaneous direct kinematics Jacobians.
%===============================================================================
\section{Conclusions}
In this paper, we presented a numerical approach to solving the Cosserat rod BVP that combines orthogonal collocation and the Magnus expansion which, when solved, results in a closed-form product of matrix exponentials equation. We have discussed the convergence of the Magnus expansion for the case of Cosserat rods and showed in simulation that both the fourth order and sixth order Magnus expansions provide accurate solutions to the BVP and that a small number of collocation points can provide reasonably accurate results. Although our results are preliminary, we believe this method has the potential to provide reduced computational cost when solving Cosserat rod models. Future work will study Jacobian-based analysis of the resulting product of exponentials and implementations for robotic applications.
%===============================================================================
\bibliographystyle{IEEEtran}
\bibliography{main}
\end{document}